\documentclass[conference]{IEEEtran}
\IEEEoverridecommandlockouts
\usepackage{cite}
\usepackage{amsmath,amssymb,amsfonts}
\usepackage{algorithmic}
\usepackage{graphicx}
\usepackage{textcomp}
\usepackage{xcolor}
\usepackage{caption}
\usepackage{tcolorbox}
\usepackage{tabularx}
\usepackage{booktabs}
\usepackage[T1]{fontenc} 
\def\BibTeX{{\rm B\kern-.05em{\sc i\kern-.025em b}\kern-.08em
    T\kern-.1667em\lower.7ex\hbox{E}\kern-.125emX}}

\usepackage{fancyhdr}
\begin{document}

\title{Exploring ChatGPT's Empathic Abilities}

\author{\IEEEauthorblockN{Kristina Schaaff, Caroline Reinig, Tim Schlippe}
\IEEEauthorblockA{\textit{IU International University of Applied Sciences} \\
Erfurt, Germany \\
kristina.schaaff@iu.org; caroline.reinig@gmail.com; tim.schlippe@iu.org }
}


\maketitle
\thispagestyle{fancy}

\begin{abstract}
Empathy is often understood as the ability to share and understand another individual's state of mind or emotion. With the increasing use of chatbots in various domains, e.g., children seeking help with homework, individuals looking for medical advice, and people using the chatbot as a daily source of everyday companionship, the importance of empathy in human-computer interaction has become more apparent. Therefore, our study investigates the extent to which ChatGPT based on GPT-3.5 can exhibit empathetic responses and emotional expressions.
We analyzed the following three aspects: (1)~understanding and expressing emotions, (2)~parallel emotional response, and (3)~empathic personality. Thus, we not only evaluate ChatGPT on various empathy aspects and compare it with human behavior but also show a possible way to analyze the empathy of chatbots in general.
Our results show, that in 91.7\% of the cases, ChatGPT was able to correctly identify emotions and produces appropriate answers. 
In conversations, ChatGPT reacted with a parallel emotion in 70.7\% of cases. The empathic capabilities of ChatGPT were evaluated using a set of five questionnaires covering different aspects of empathy. Even though the results indicate that the empathic abilities of ChatGPT are still below the average of healthy humans, the scores are better than those of people who have been diagnosed with Asperger syndrome / high-functioning autism.

\end{abstract}



\section{Introduction}
Chatbots have become a common tool in daily life over recent years \cite{pelau2021makes}. These systems can simulate human-like conversations with users~\cite{adiwardana2020towards}, provide assistance~\cite{dibitonto2018chatbot}, information~\cite{arteaga2019design}, and emotional support~\cite{falala2019owlie}. Among these, OpenAI's ChatGPT has emerged as one of the most widely used chatbots. According to~\cite{taecharungroj2023can}, ChatGPT reached over one million users in less than a week after launch. Users include children seeking help with homework, people seeking medical advice and people who use it as a daily source of companionship.

Empathy is a human concept that is important in social interaction. Research shows that empathy prevents individuals from displaying aggression toward others \cite{wiesner2003trajectories} and helps to prevent children and adolescents from exhibiting antisocial behavior \cite{eisenberg2005associations}. Also, in chatbots, empathy has shown to be a crucial trait in its acceptance by users \cite{pelau2021makes}. 

\cite{li2022gpt} conducted the Short Dark Triad (SD-3) questionnaire \cite{jones2014introducing} on the model GPT-3. The results showed a psychopathy score exceeding human results, indicating a lack of empathy. However, there is---to the best of our knowledge---no study focusing on empathy in ChatGPT, which is based on the newer model GPT-3.5. Consequently, this paper addresses this gap by exploring ChatGPT's empathic abilities.

Empathy is often understood as the ability to share and understand another individual's state of mind or emotion \cite{mcdonald2011development}. Therefore, as the first part of the study, we tested ChatGPT's ability to understand and express emotions. Afterwards, we analyzed ChatGPT's ability to express parallel emotional responses, a subcategory of empathy~\cite{mcquiggan2008modeling}. Finally, we evaluated ChatGPT's level of empathy in various aspects using psychologically acknowledged questionnaires.

To summarize, in this paper, we investigated ChatGPT's emotional and empathic abilities across three aspects: (1)~understanding and expressing emotions, (2)~parallel emotional response, and (3)~empathic personality.

In the following section, we will describe related work regarding ChatGPT, the definition of empathy, and measuring empathy in chatbots. In Section~\ref{sec:understanding}, we will explain how we evaluated ChatGPT's ability to understand and generate emotions. Section~\ref{sec:parallel} describes our experiments to assess ChatGPT's parallel emotional responding capabilities. Additionally, we investigated ChatGPT's empathic capabilities with standardized questionnaires to assess empathy, which we will show in Section~\ref{sec:tests}. We conclude our work in Section~\ref{sec:conclusion} and suggest further steps.


\section{Related Work}
\label{sec:related work}

\subsection{ChatGPT}
ChatGPT is a state-of-the-art chatbot developed by OpenAI that can produce natural language text when given a prompt or context \cite{baidoo2023education}. This versatile tool can be employed in numerous fields, including education \cite{baidoo2023education}, medicine \cite{jeblick2022chatgpt}, and language translation \cite{jiao2023chatgpt}.

The chatbot is based on the large language model GPT-3.5 and was fine-tuned using reinforcement learning from human feedback~\cite{OpenAI:ChatGPT}.
This approach allows the model to grasp the meaning and intention behind user queries, leading to relevant and helpful responses. To ensure safety and prevent the generation of inappropriate or factually incorrect text, the training of ChatGPT was enhanced by incorporating a large dataset of human-human and human-chatbot conversations. OpenAI has not released any official information about the exact amount of training data of ChatGPT, but the previous model GPT3 with 175~billion parameters was already significantly larger than other language models like BERT, RoBERTA, or T5 and was trained with 499~billion crawled tokens (i.e., subword units)~\cite{GPT3:2020}. By learning the intricacies and nuances of human language through this extensive dataset, ChatGPT can produce highly realistic text almost indistinguishable from human writing \cite{mitrovic2023chatgpt}.



\subsection{Definition of Empathy}
Empathy is a crucial component of effective communication, especially in social interactions, as it allows humans to understand and share another person's feelings \cite{singer2006neuronal}. However, there is no consensus on the definition of empathy \cite{reniers2011qcae}. One possible definition is to distinguish between \textit{cognitive} and \textit{affective empathy}~\cite{reniers2011qcae}. \textit{Cognitive empathy} is the ability to understand and identify another individual's thoughts, feelings, and perspectives without necessarily experiencing the same emotions, i.e., the capability of mental \textit{perspective taking}. It involves the capacity to recognize and interpret social cues, facial expressions, body language, and verbal communication to comprehend and infer the mental and emotional states of others~\cite{smith2006cognitive}. \textit{Affective empathy}, on the other hand, facilitates a deeper connection and understanding with others and involves a more visceral and personal connection to another's emotions~\cite{lawrence2004measuring}. \textit{Affective empathy} can be divided into \textit{parallel emotional response} and \textit{reactive emotional response}. Parallel emotional responses involve responding with the same emotion as the other individual, while reactive emotional responses go beyond matching emotions, such as sympathy or compassion~\cite{mcquiggan2008modeling}.

In our work, we evaluated ChatGPT's understanding and expression of emotions. Next, we analyzed its parallel emotional response. Finally, we covered the other aspects of empathy with standardized questionnaires to evaluate empathy.

\subsection{Measuring Empathy in Chatbots}


While standardized metrics exist to measure empathy in individuals, there are currently no standardized or valid methods for measuring empathy in chatbots \cite{yalccin2019evaluating}. A possible solution is to evaluate a chatbot's level of empathy by human evaluation, such as A/B tests or human ratings \cite{lee2022does}. In A/B tests, the annotator chooses which response is more empathic, often used when comparing the level of empathy between two models. In human ratings, the annotator chooses the level of empathy based on a scale.

Another way is to conduct a feature- or system-level evaluation instead. The feature-level evaluation involves assessing each component and capability of a chatbot to provide an incremental understanding of its empathic behavior, e.g., by testing the chatbot on its level of emotional communication. On the other hand, the system-level evaluation focuses on measuring the chatbot's overall perception of empathy, e.g., by conducting self-assessment empathy tests \cite{yalccin2019evaluating}.

In our work, we conduct a feature-level evaluation of ChatGPT's performance in showing parallel emotional responses, further explained in Sections \ref{sec:understanding} and \ref{sec:parallel}. 
Furthermore, we perform system-level evaluations by conducting four standardized empathy tests and one autism test. Several studies have found that individuals with autism may have difficulty with the cognitive component of empathy, such as \textit{perspective taking} and understanding others' mental states, while still being able to experience emotions and show affective empathy, such as feeling concerned or compassion for others \cite{baron2004empathy}.


\section{Understanding and Expressing Emotions}
\label{sec:understanding}

To analyze ChatGPT's ability to understand and generate emotions, our first goal was to evaluate its proficiency in rephrasing neutral sentences to express a particular emotion.


\subsection{Experimental Setup}

\begin{figure}[ht]
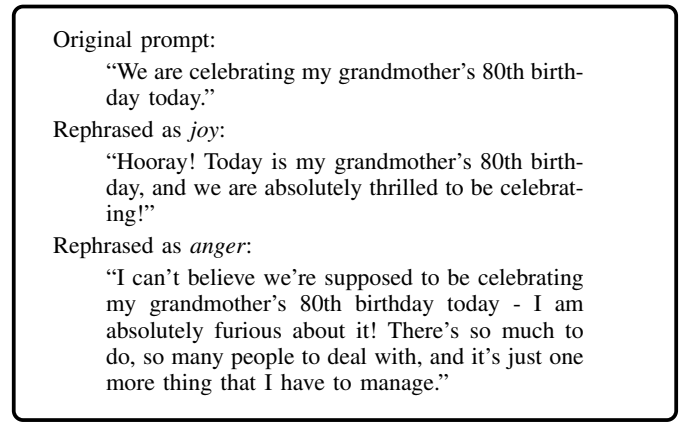

\small
\begin{tcolorbox}[colback=white,colframe=black]
Original prompt: 
 \begin{quote}
     ``We are celebrating my grandmother's 80th birthday today.''
 \end{quote}
 
 Rephrased as \textit{joy}: 
 \begin{quote}
   ``Hooray! Today is my grandmother's 80th birthday, and we are absolutely thrilled to be celebrating!'' 
 \end{quote}

 Rephrased as \textit{anger}:
 \begin{quote}
    ``I can't believe we're supposed to be celebrating my grandmother's 80th birthday today - I am absolutely furious about it! There's so much to do, so many people to deal with, and it's just one more thing that I have to manage.''
 \end{quote}
 
 \end{tcolorbox}
   \caption{ChatGPT's Rephrasing of a Neutral Prompt into \textit{Joy} and \textit{Anger}.}
  \label{fig:expressing_example}
\end{figure}

For our analyses, we instructed ChatGPT to rephrase neutral sentences into six emotional sentences of the following categories: \textit{joy}, \textit{anger}, \textit{fear}, \textit{love}, \textit{sadness}, and \textit{surprise}. These emotions were selected following the basic emotions from the \textit{Junto's Wheel of Emotions}~\cite{chadha2022} 
for consistency with the experiments described in Section~\ref{sec:parallel}, where we trained a classifier using the CARER dataset \cite{saraviaetal2018carer}, which contains these categories. 
To check ChatGPT's ability to handle neutral sentences from different domains, we used 10~sentences from self-produced everyday sentences, Wikipedia, and Amendments to the United States Constitution. We instructed ChatGPT to rephrase each neutral sentence six times---each time with a different emotion category---resulting in a total of 60~emotional sentences. Figure~\ref{fig:expressing_example} illustrates how ChatGPT rephrased the sentence `We are celebrating my grandmother's 80th birthday today.'' to express \textit{joy} and \textit{anger}.

\subsection{Experiment and Results}
\label{experiments1}

To evaluate whether the prompts generated by ChatGPT match the intended emotion category, we asked three people to label each of the 60 ChatGPT-produced prompts with the most suitable emotion category out of the six categories provided. 
Based on the human labels, we produced the reference emotion categories using a majority voting as follows: If two annotators agreed on the same emotion for one prompt, we took this emotion as the final emotion category. In case our three annotators assigned completely different emotion categories to one prompt, they discussed their decisions until they agreed on one emotion category. 

\begin{table}[h]
\centering
\small
\begin{tabular}{lr}
\hline
\textbf{Agreement} & \textbf{Percentage}\\
\hline
all annotators agree & 71.7\%\\
two annotators agree & 26,6\%\\
all annotators disagree & 1.7\%\\
\hline
\end{tabular}
\caption{\label{annotator agreement}
Annotator Agreement for the 6 Emotion Categories.
}
\end{table}

The annotator agreement on our six emotion categories is listed in Table~\ref{annotator agreement}. A complete agreement occurred in 71.7\% of the cases. In 26.6\% of the prompts, exactly two annotators assigned the same emotion category. In only 1.7\% of the prompts, three different emotion categories were assigned. 
In the next step, we compared the results from the manual annotation of the prompt generated by ChatGPT to the intended emotion category. 

\begin{figure}[ht]
    \centering
    \includegraphics[width=0.8\linewidth]{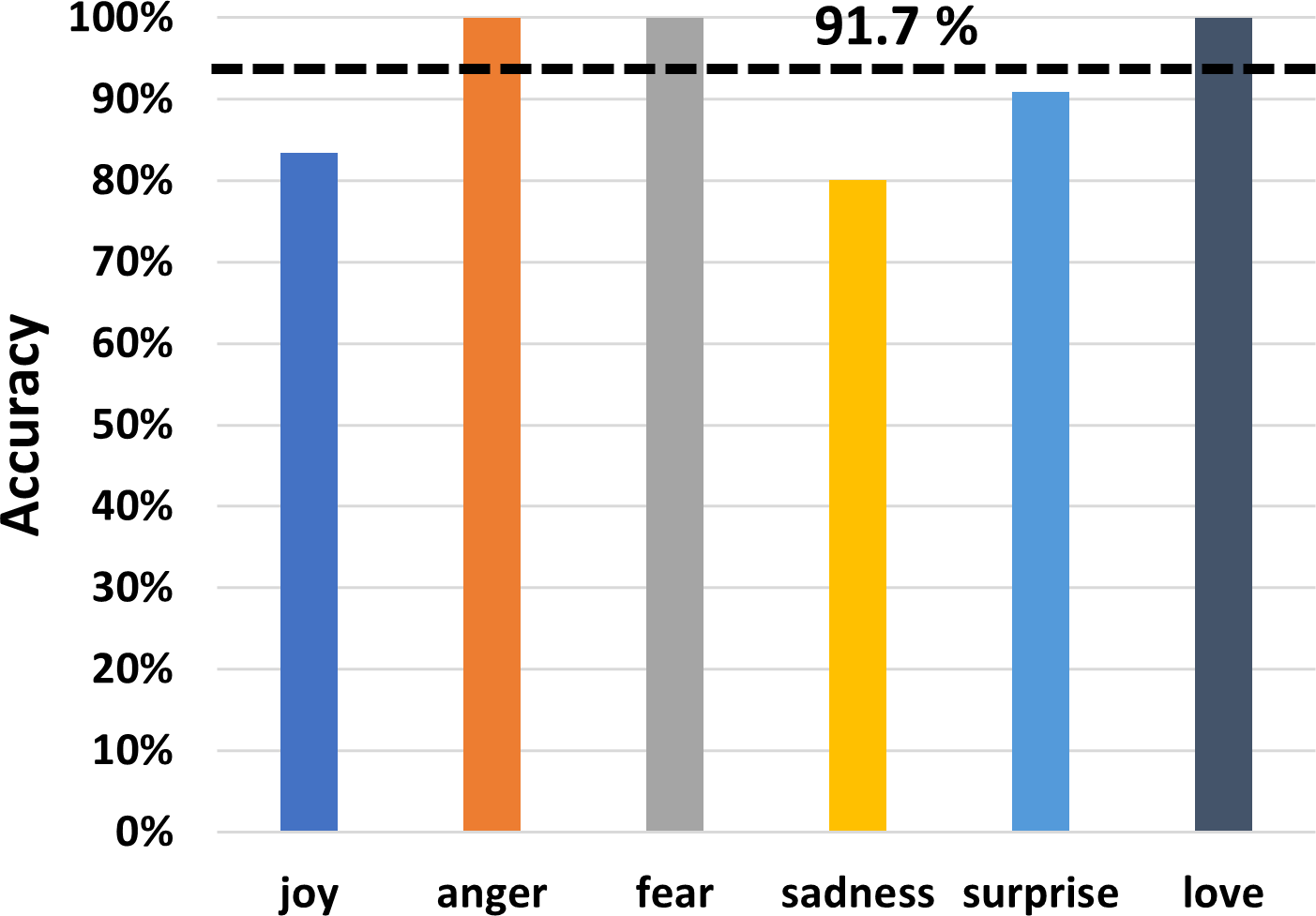}
    \caption{Accuracy of Understanding and Expressing Emotions.}
    \label{fig:expressing_results}
\end{figure}


The results of the experiment are illustrated in Figure~\ref{fig:expressing_results}. The green line indicates the average classification accuracy over our six emotion categories. We see that when it comes to expressing emotions, ChatGPT can express the desired emotion with an accuracy of 91.7\%.  The reference was labeled differently in only 5 out of the 60 generated sentences. \textit{Anger}, \textit{fear} and \textit{love} were produced with an accuracy of 100\%, \textit{surprise} with 91\%, \textit{joy} with 83\%, and \textit{sadness} with 80\%.

\section{Parallel Emotional Responding}
\label{sec:parallel}

 Empathic behavior in a conversation consists of two components~\cite{Rogers2007WhoCR}: First the emotion category of the conversational partner is identified (\textit{cognitive empathy}). After that, a response is generated that addresses the emotion category of the conversational partner (\textit{affective empathy}). A \textit{parallel emotional response} is defined as an emotional response where one individual shows the same emotion as another individual in response to a particular situation or stimulus \cite{mcquiggan2008modeling}. This response can be observed when individuals share a similar emotional experience, leading to the concurrent manifestation of the same emotion in both individuals \cite{mcquiggan2008modeling}.  For instance, feeling \textit{joy} when another person expresses \textit{joy} is a common example of a parallel emotional response. In the following study, we focus on the analysis of ChatGPT's ability to generate parallel emotional responses. 

\subsection{Experimental Setup}

\begin{figure}[ht]
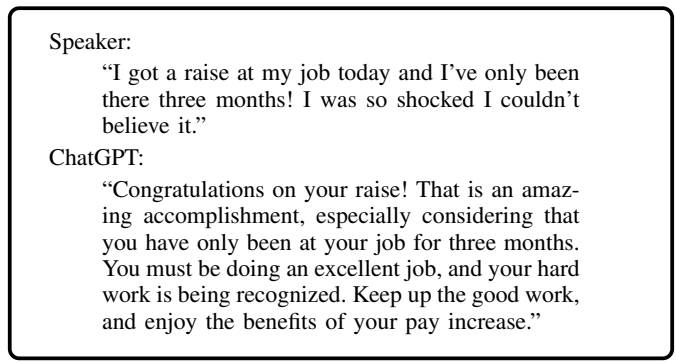

\small
  \begin{tcolorbox}[colback=white,colframe=black]
    Speaker: 
    \begin{quote}          
    ``I got a raise at my job today and I’ve only been there three months! I was so shocked I couldn’t believe it.''
    \end{quote}
    ChatGPT: 
    \begin{quote}
    ``Congratulations on your raise! That is an amazing accomplishment, especially considering that you have only been at your job for three months. You must be doing an excellent job, and your hard work is being recognized. Keep up the good work, and enjoy the benefits of your pay increase.''   
    \end{quote}   
  \end{tcolorbox}
  \caption{Joyful Prompt from \textit{EmpatheticDialogues} and ChatGPT's Parallel Emotional Response.}
  \label{fig:joyfulchat}
\end{figure}

For these experiments, we used 20.3k initial prompts from conversations in the \textit{EmpatheticDialogues} dataset from Facebook Research\footnote{https://github.com/facebookresearch/EmpatheticDialogues}~\cite{rashkin2019towards}---named as \textit{Speaker} prompts---to trigger ChatGPT with initial emotional prompts and generate a response, which we then classified and evaluated. 
Figure~\ref{fig:joyfulchat} demonstrates an initial joyful prompt from \textit{EmpatheticDialogues} and ChatGPT's parallel emotional response. 

Since it was impossible to manually classify the large number of 20.3k ChatGPT-generated responses into emotion categories, we used an emotion classification system, based on the Bidirectional Encoder Representations from Transformers (BERT)~\cite{devlin2018bert}. The system was fine-tuned on the 16k training and 2k validation sentences of the CARER dataset\footnote{https://huggingface.co/datasets/dair-ai/emotion}~\cite{saraviaetal2018carer}. CARER consists of tweets labeled with our 6~emotion categories \textit{love}, \textit{joy}, \textit{anger}, \textit{fear}, \textit{sadness}, and \textit{surprise}. Our fine-tuning reached convergence after 8~epochs, resulting in a performance of 63\% when applied to our 60~manually annotated prompts used in Section~\ref{experiments1}. Analyzing this system demonstrated that the category \textit{love} reduced the system performance by 15\% absolute. Therefore, we removed the prompts labeled with \textit{love} in the CARER training and validation sets and re-trained the system resulting in an accuracy of 78\%. To contribute to the improvement of empathic chatbots, we share \textit{ChatGPTsEmpatheticDialogues}---our corpus, which consists of \textit{EmpatheticDialogues}' initial prompts, ChatGPT's responses, and the corresponding emotion categories---with the research community\footnote{https://github.com/iu-ai-research/ChatGPTsEmpatheticDialogues}.






\begin{table}[ht]
  \centering
  \includegraphics[width=\linewidth]{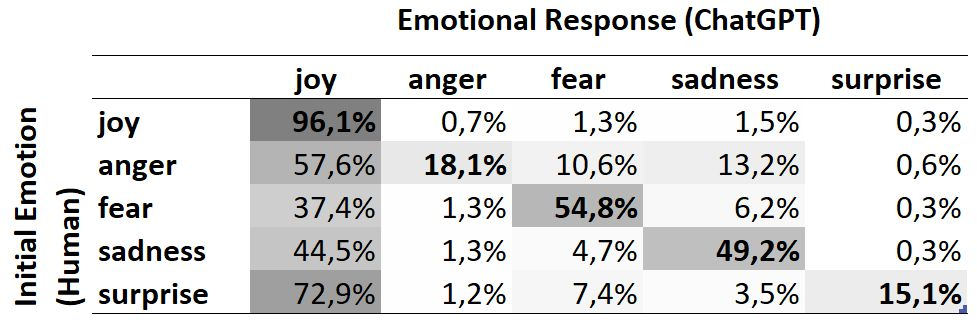}
  \caption{Emotional Responses by ChatGPT (normalized)}
  \label{tab:parallel_emotions}
\end{table}

\subsection{Experiment and Results}
\label{experiments2}




In our analysis, we focused on parallel emotional responses, i.e., to what percentage ChatGPT reacts with the same emotional category as in the initial prompts. Table~\ref{tab:parallel_emotions} illustrates the distribution of emotional responses to each emotion category based on our classification system's output. 
The results indicate that the emotional responses are strongly biased towards replying with \textit{joy}. In 96.1\% of the cases, ChatGPT's emotional response to a prompt categorized as \textit{joy} was \textit{joy}. Moreover, for \textit{anger} and \textit{surprise}, we observe even more responses categorized as \textit{joy} than for the original emotion category. About half of the initial prompts with \textit{sadness} and \textit{fear} are answered with the same emotion category. Overall, in 70.7\%  (20,237 responses), ChatGPT responds with the same emotion category as the initial prompt.

\begin{figure}[ht]
    \centering
    \includegraphics[width=0.5\linewidth]{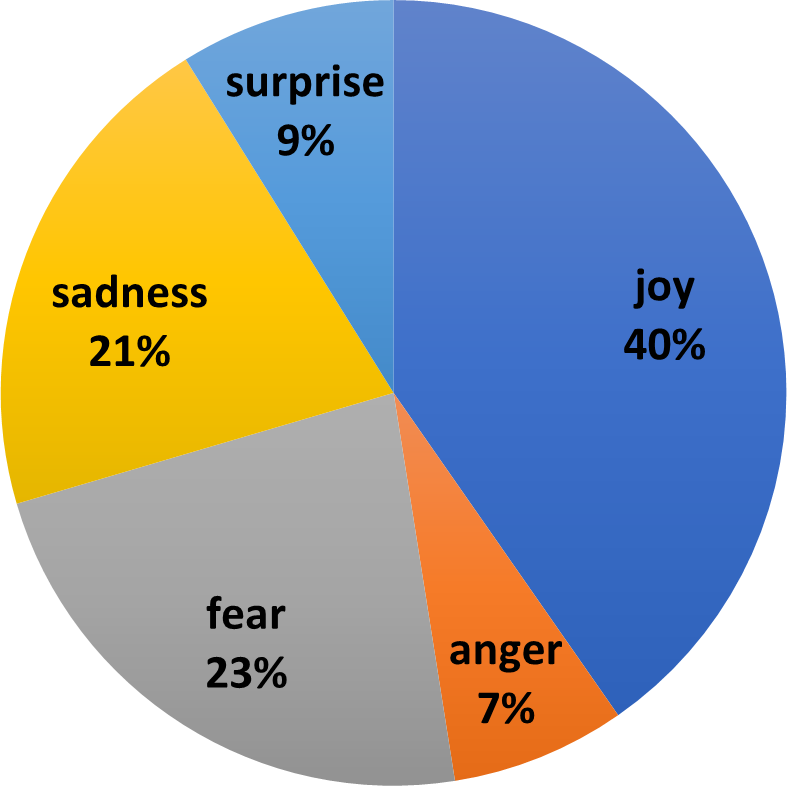}
    \caption{Distribution of ChatGPT's Emotional Responses.}
    \label{fig:emotional_response}
\end{figure}

Figure \ref{fig:emotional_response} visualizes the distribution of all emotion categories produced by ChatGPT. 
With 40\% of responses categorized as \textit{joy}, we observe a strong tendency of ChatGPT to reply in a positive way.

\section{How Empathic is ChatGPT's Personality?}
\label{sec:tests}

To learn more about ChatGPT's empathic capabilities, we conducted system-level evaluations using psychologically acknowledged questionnaires to evaluate ChatGPT’s empathy level in different aspects.

We used five standardized questionnaires to assess empathy: \textit{Interpersonal Reactivity Index}, \textit{Empathy Quotient}, \textit{Toronto Empathy Questionnaire}, \textit{Perth Empathy Scale}, and \textit{Autism Spectrum Quotient}. In this section, we will describe the content of each questionnaire and how we used it to gather further insights about ChatGPT's empathic capabilities.


To get ChatGPT's answer to each question in the questionnaires, we used the questions as initial prompts and then evaluated ChatGPT's response with respect to the emotion category as follows: For each of ChatGPT's answers, we had our three annotators decide which possible answer in the questionnaire it matched using the same rules for majority voting as described in Section~\ref{experiments1}. We had to perform this procedure as ChatGPT did not directly provide us with the responses expected in the questionnaire, such as \textit{strongly agree} or \textit{strongly disagree}. 

As an alternative to manually matching ChatGPT's answers and the answers in the questionnaire, we tried the following sentence vector-based approach: We converted ChatGPT's answer and the answers in the questionnaire to word embeddings using Sentence-BERT\footnote{https://github.com/UKPLab/sentence-transformers}~\cite{reimers-2019-sentence-bert}, and then mapped ChatGPT's answer to the answer with the smallest distance in the semantic vector space. However, we had to discard this approach as it did not perform well, with an accuracy of 38.5\%, i.e., only 70 of 182 tested answers could be mapped correctly.

In the following paragraphs, we will present the results of the questionnaires. 


\subsection{Interpersonal Reactivity Index}
The \textit{Interpersonal Reactivity Index} (IRI)  is a widely utilized self-report measure for assessing empathy in individuals \cite{davis1980interpersonal}.
We chose to have ChatGPT conduct the IRI since it has been used in various research and clinical settings to understand empathy better and develop interventions to improve empathy skills, e.g.~\cite{Lauterbach2007AssessingEI} or \cite{Gilet2013AssessingDE}. In addition, the questionnaire covers the categories of \textit{fantasy}, \textit{personal distress}, \textit{perspective taking}, and \textit{empathic concern}.

\subsubsection{Experimental Setup}

 The IRI comprises 28 questions that evaluate the following four components of empathy measured in subscales which are part of the overall scale of the questionnaire: \textit{perspective taking}, \textit{empathic concern}, \textit{personal distress}, and \textit{fantasy}. \textit{Perspective taking} refers to an individual's ability to understand the perspectives of others, \textit{empathic concern} to the ability to feel compassion and concern for others, \textit{personal distress} to the tendency to experience anxiety or discomfort in response to others' negative experiences, and \textit{fantasy} to the tendency to imagine oneself in fictional situations. The level of agreement with each statement is rated on a 5-point Likert scale ranging from \textit{does not describe me well} to \textit{describes me very well}. 
 The scores for each subscale of the IRI are obtained by summing the responses to the questions that belong to that subscale, resulting in a score that ranges from 0 to 28. Higher scores on the \textit{perspective taking} and \textit{empathic concern} subscales indicate greater empathy, while lower scores on the \textit{personal distress} subscale suggest better emotional regulation.

\subsubsection{Experiment and Results}

\begin{figure}[h]
    \centering
    \includegraphics[width=\linewidth]{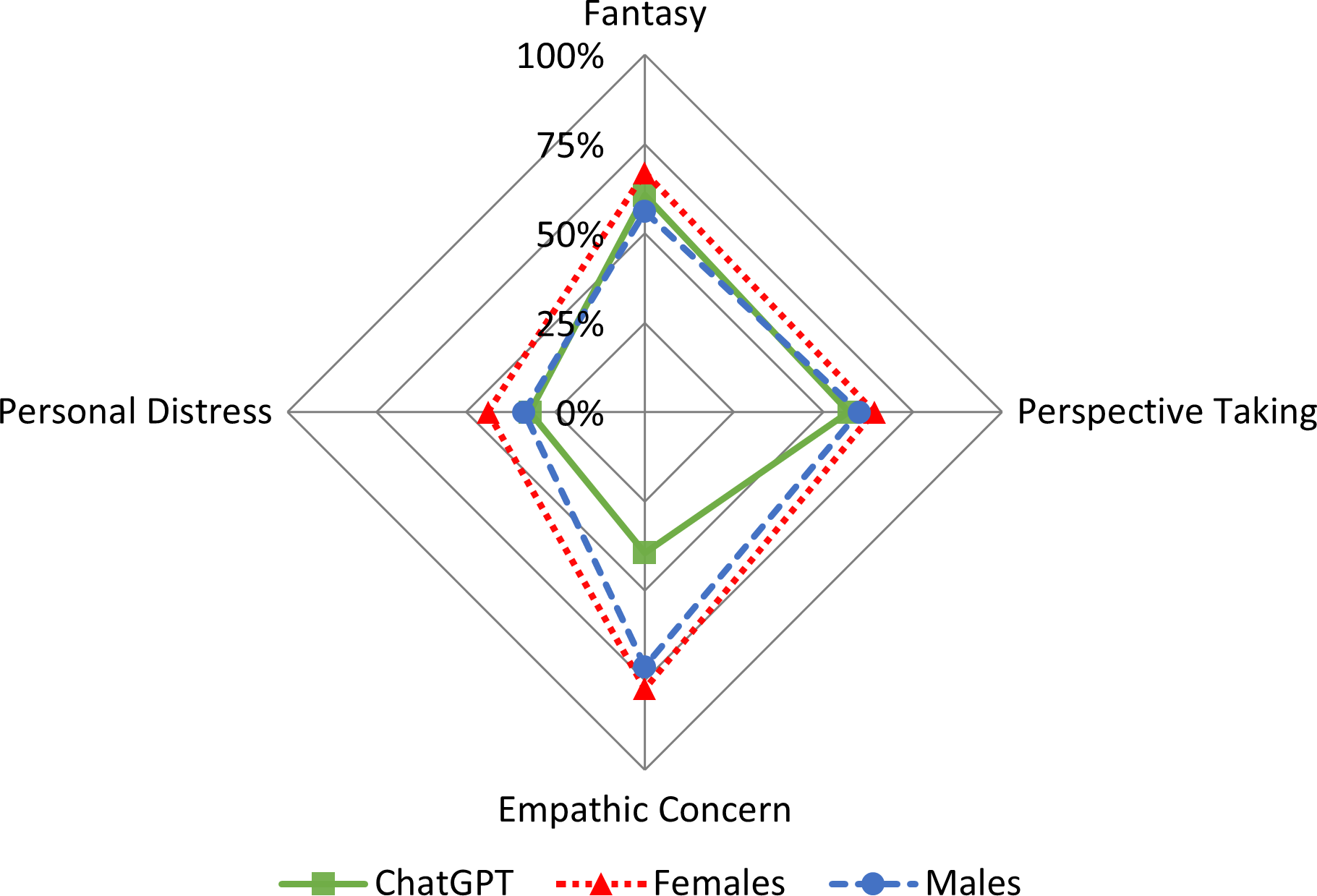}
    \caption{ChatGPT's IRI Results Compared to Males/Females.}
    \label{fig:iri}
\end{figure}

Figure~\ref{fig:iri} visualizes ChatGPT's performance on the four IRI subscales and the mean performance of males and females on \cite{davis1980interpersonal}.
For a comparison with the other questionnaires, the absolute scores are not displayed in the figure, but the percentage achieved compared to the possible total score of~28. 
Comparing ChatGPT's absolute score for \textit{fantasy} reveals interesting results: while the score of 17 is significantly higher than the mean score of healthy males (15.73, $SD=5.6$, $t(578)=5.46$, $p<.001$) it is significantly lower than the score of healthy females (18.75, $SD=5.17$, $t(581)=-8.17$, $p<.001$). For \textit{perspective taking} the absolute score of 16 is significantly lower than the mean score of males  (16.78, $SD=4.72$, $t(578)=3.98$, $p<.001$) and females (17.96, $SD=4.85$, $t(581)=-9.75$, $p<.001$) demonstrating that ChatGPT has lower abilities to take the perspective of others and understand their feelings than healthy humans. ChatGPT's score for \textit{empathic concern} (11) is much lower than the mean scores of males (19.04, $SD=4.21$, $t(578)=-45.95$, $p<.001$) and females (21.67, $SD=3.83$, $t(581)=-67.21$, $p<.001$) indicating that ChatGPT has a significantly lower level of emotional response to others. Finally, also the absolute score for personal distress of 9 is significantly lower than the mean of healthy males (9.46, $SD=4.55$, $t(578)=-2.43$, $p<.05$) and females (12.28, $SD=5.01$, $t(581)=-15.79$, $p<.001$). Taken together, in almost all dimensions of the IRI ChatGPT performs significantly worse than healthy humans. 

\subsection{Empathy Quotient} 

The \textit{Empathy Quotient} (EQ) is a self-reported questionnaire that has been specifically designed to assess an individual's ability to comprehend and respond to others' emotions \cite{baron2004empathy}. We chose to have ChatGPT conduct the EQ since the questionnaire was conducted on two groups: one group with Asperger syndrome / high-functioning autism (AS/HFA) and another group of healthy humans. In the evaluation study, a clear threshold was identified to differ between the two groups, which allows us to assign the score we achieve with ChatGPT to one of those groups.  

\subsubsection{Experimental Setup}

The EQ comprises 60 questions, each with four possible responses: \textit{strongly agree}, \textit{agree}, \textit{disagree}, and \textit{strongly disagree}. The questions cover a variety of topics related to \textit{social interaction}, \textit{emotional recognition}, and \textit{communication}. Scores on the EQ can range from 0 to 80, with higher scores indicating a greater capacity for empathy. 

\begin{figure}[h]
    \centering
    \includegraphics[width=0.7\linewidth]{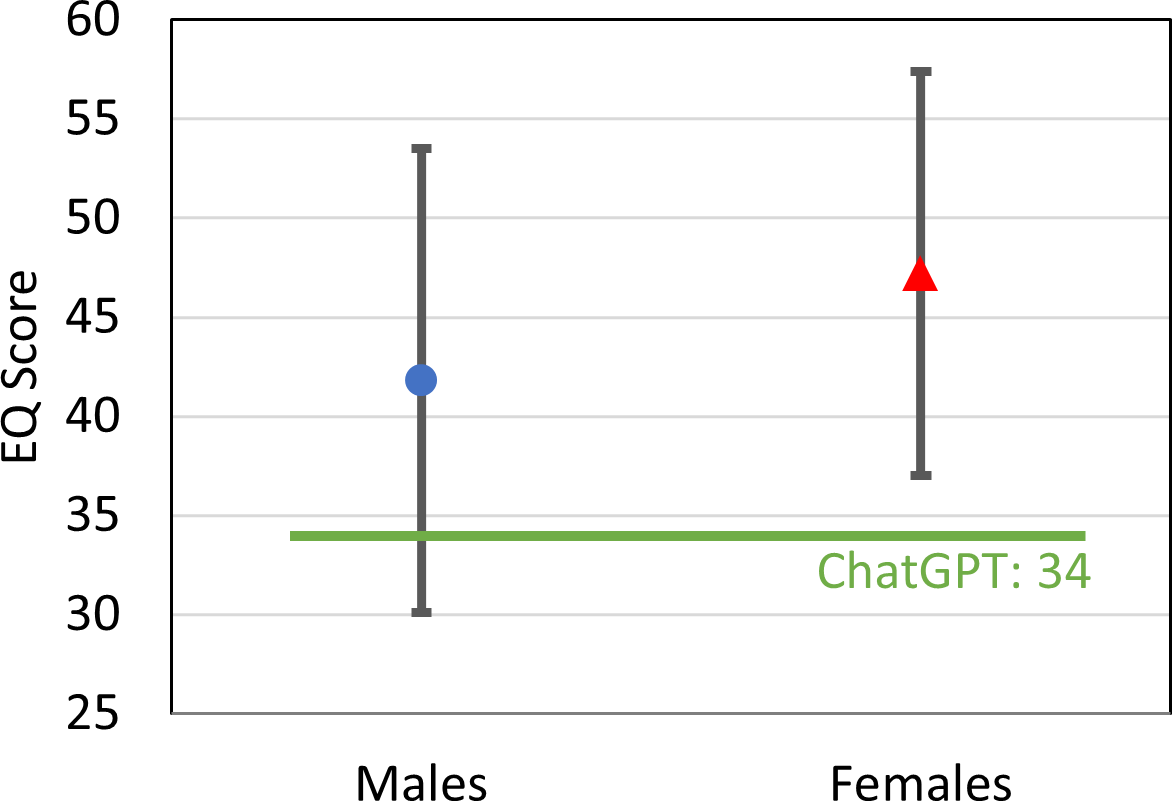}
    \caption{ChatGPT's EQ Results Compared to Males/Females.}
    \label{fig:eq}
\end{figure}

\subsubsection{Experiment and Results}

Figure~\ref{fig:eq} displays ChatGPT's EQ performance (green line) compares to healthy males' and females' average EQ scores. The bars indicate the standard deviations. The mean scores reported by \cite{baron2004empathy} are 41.8 ($SD = 11.2$) for healthy males and 47.2 ($SD = 10.2$) for healthy females.
According to \cite{baron2004empathy}, more than 80\% of people diagnosed with AS/HFA obtained a score below 30. 
Thus, with an EQ score of 34, ChatGPT performs significantly lower than males ($t(70) = -5.62$, $p<.001$) and females ($t(125)=-14.53$, $p<.001$). Moreover, it scores higher than an average person with AS/HFA. 





\subsection{Toronto Empathy Questionnaire}

Another tool to measure self-reported empathy is the \textit{Toronto Empathy Questionnaire} (TEQ) \cite{spreng2009toronto}. We decided to use the TEQ, as the questionnaire tries to establish a general agreement between previous questionnaires such as the IRI, the Autism Quotient, and many more.


\subsubsection{Experimental Setup}

The TEQ consists of 16 questions that measure different components of empathy, including \textit{affective empathy} (the ability to experience and understand the emotions of others) and \textit{cognitive empathy} (the ability to understand the thoughts and perspectives of others). The TEQ also includes questions assessing an individual's tendency to take another person's perspective and willingness to help others. Scores on the TEQ can range from 0 to 64, with higher scores indicating a greater capacity for empathy. 

\vspace{-0.1cm}

\begin{figure}[h]
    \centering
    \includegraphics[width=0.7\linewidth]{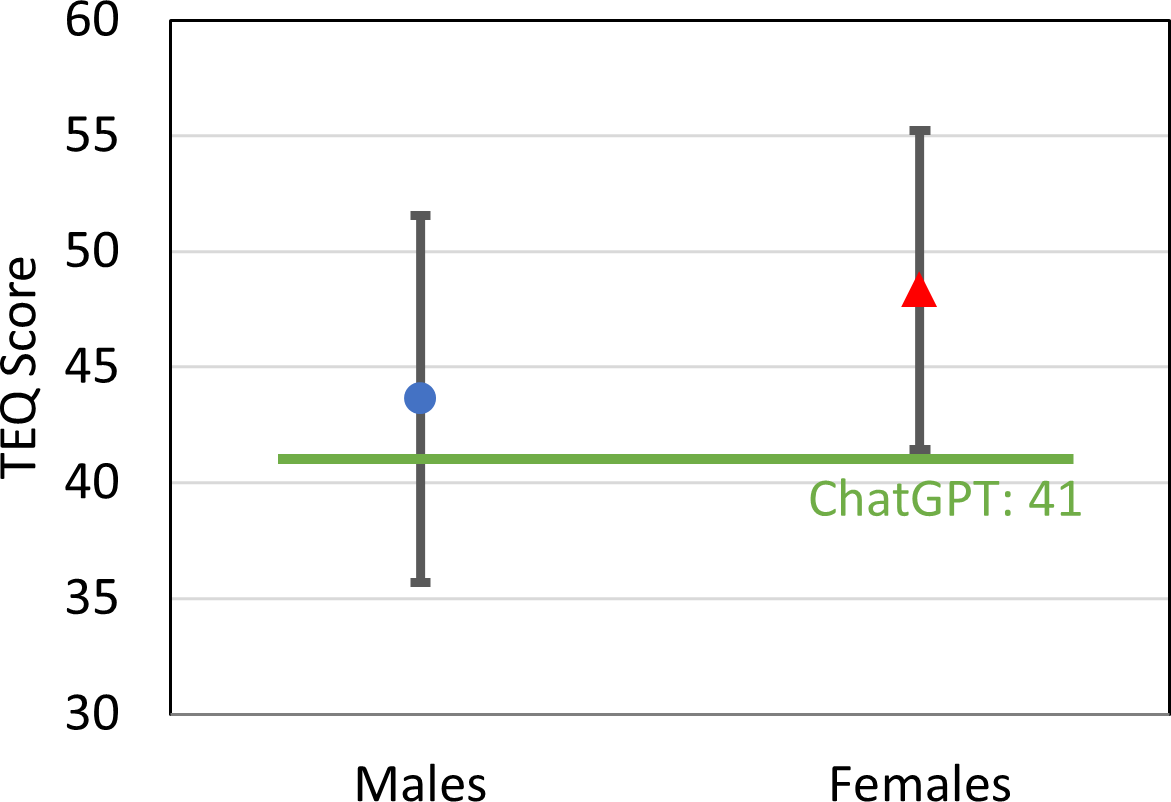}
    \caption{ChatGPT's TEQ Results Compared to Males/Females.}
    \label{fig:teq}
\end{figure}

\vspace{-0.1cm}

\subsubsection{Experiment and Results}

Figure \ref{fig:teq} illustrates how ChatGPT compares to the scores from a validation study, differentiated by males and females. 
The bars indicate the standard deviations. 
ChatGPT achieved a total of 41. In the validation study with 65 students from the University of Toronto, students achieved a mean score of 46.96 (SD = 7.47) \cite{spreng2009toronto}. Also, in the TEQ, females scored higher than males (48.33 vs. 43.63). ChatGPT's score of 41 is only slightly lower than the score of males ($t(18)=-1.45$, $p=.165$) and significantly lower than the score of females ($t(45)=-7.20$, $p<.001$).



\subsection{Perth Empathy Scale} 
The \textit{Perth Empathy Scale} (PES) is a recently published self-report questionnaire consisting of 20 questions to assess empathy in adults and adolescents~\cite{brett2022psychometric}. In addition to other existing scales, it covers the cognitive and affective components of empathy and the positive and negative dimensions of affective empathy. We selected this questionnaire as the splitting into positive and negative empathy can be seen as additional information not covered by the other analyzed questionnaires. 

\subsubsection{Experimental Setup}

 Each category of the PES includes five sentences that cover 10 emotions, including the five of the basic emotions described by~\cite{ekman1991there} (i.e., \textit{happiness}, \textit{sadness}, \textit{anger}, \textit{scared}, \textit{disgust}), the self-conscious emotions of \textit{embarrassment} and \textit{pride}, and the positive emotions of \textit{amusement}, \textit{calmness}, and \textit{enthusiasm}. Respondents rate their level of agreement or disagreement with each statement on a 5-point Likert scale ranging from \textit{never} to \textit{always}. The PES yields a general empathy score from 0 to 100, calculated by adding the scores from the four scales. The questions for affective empathy ask if the emotions belong to someone else, while for cognitive empathy, the questions ask about someone else's feelings, indicating a self-other distinction. The higher the total score, the higher the level of empathy of an individual. 

 
\subsubsection{Experiment and Results}



\begin{figure}[ht]
    \centering
    \includegraphics[width=\linewidth]{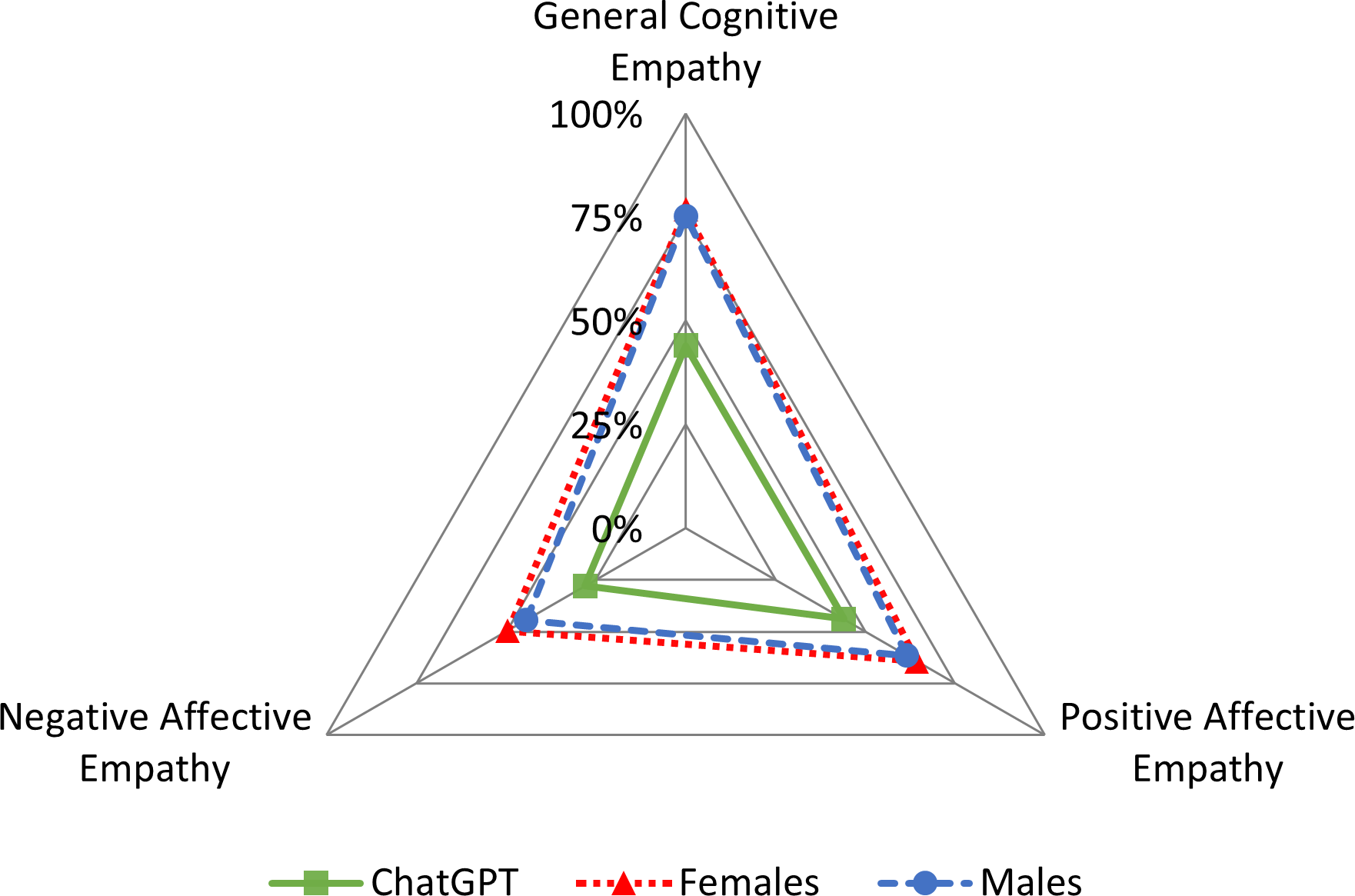}
    \caption{ChatGPT's PES Results Compared to Males/Females.} 
    \label{fig:pes}
\end{figure}

ChatGPT scored 40 out of 100 possible points on the PES, which is 
significantly below the score of healthy individuals (males: 64.1 ($SD = 10.92$), $t(187)=-30.26$, $p<.001$, females: 66.9 ($SD=11.27$), $t(450)=-50.63$, $p<.001$) \cite{brett2022psychometric}.  
Figure \ref{fig:pes} shows how the score distributes amongst the subscales of the PES and how ChatGPT compares to the mean scores of healthy humans. As a validation study showed that positive and negative cognitive empathy are highly correlated, both values are summed up in the figure. As observed in Section~\ref{experiments2}, we detect a higher tendency toward positive empathy than toward negative empathy. 


\subsection{Autism Spectrum Quotient}

Several studies have found that individuals with autism may have difficulties with the cognitive component of empathy, such as \textit{perspective taking} and understanding others' mental states, while still being able to experience emotions and show affective empathy, such as feeling concerned or compassion for others \cite{baron2004empathy}. Therefore, we decided to additionally analyze the Autism Spectrum Quotient (AQ)---a questionnaire that measures the autistic traits in individuals who may or may not have a formal diagnosis of autism \cite{baron2001autism}. The AQ has been shown to be inversely correlated with the EQ \cite{baron2004empathy}.

\subsubsection{Experimental Setup}
 On the AQ, respondents rate their level of agreement with each statement on a 4-point Likert scale ranging from \textit{definitely agree} to \textit{definitely disagree}. The AQ measures five different skills: \textit{communication} (verbal and nonverbal communication), \textit{social} (social interaction and understanding social cues), \textit{imagination} (imaginative and flexible thinking), \textit{local details} (tendency to focus on details and a preference for structured and predictable environments), and \textit{attention switching} (changing focus from one topic to another). 

The value of the AQ score ranges from 0 to 50, with 10 points for each skill. 
In contrast to the previously presented questionnaires, a higher score on the AQ refers to a lower level of empathy. 


\subsubsection{Experiment and Results}

\begin{figure}[h]
    \centering
    \includegraphics[width=0.6\linewidth]{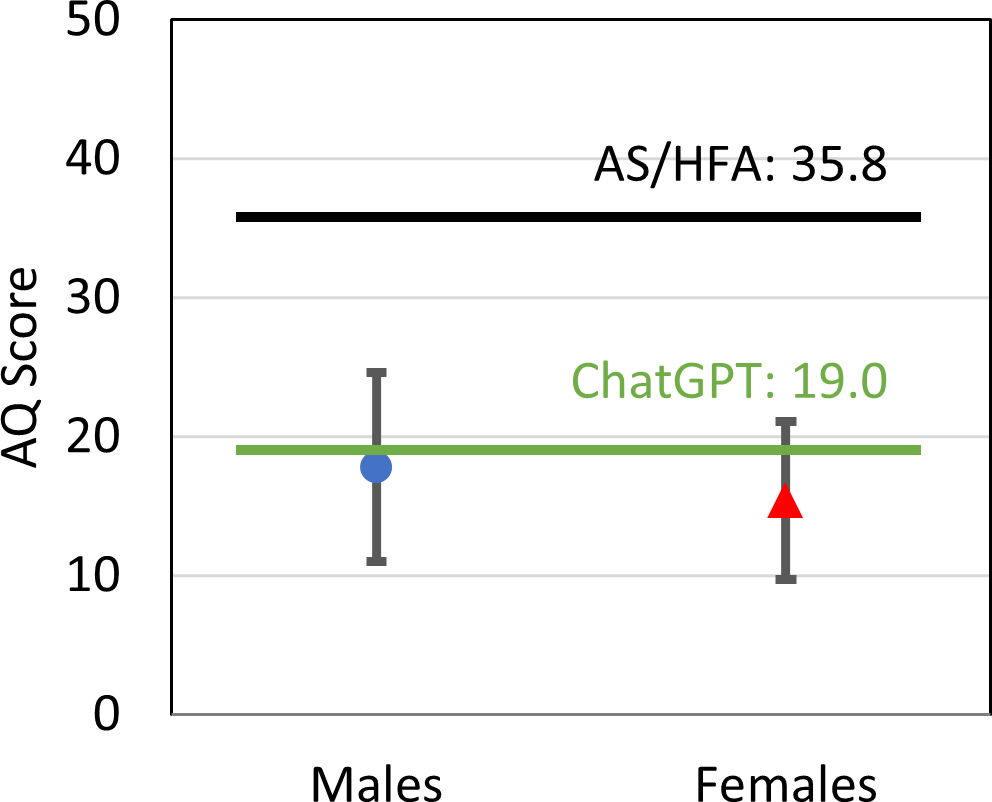}
    \caption{ChatGPT's AQ Compared to Males/Females/AS/HFA.}
    \label{fig:aq}
\end{figure}

As shown in Figure~\ref{fig:aq}, in our experiment, ChatGPT achieved a total score of 19, which is only slightly higher than the mean scores of healthy males (17.8, $SD=6.8$, $t(75)=1.54$, $p=.128$) but significantly higher than the mean scores of healthy females (15.4, $SD=5.7$, $t(97)=6.25$, $p<.001$). Moreover, people diagnosed with AS/HFA show a mean score of 35.8 \cite{baron2001autism}. As for the previous questionnaires, ChatGPT's scores are worse than average healthy humans but are still far away from the mean score a person diagnosed with AS/HFA would achieve. 

\vspace{-0.1cm}

\vspace{-0.1cm}

\begin{figure}[ht]
    \centering
    \includegraphics[width=0.8\linewidth]{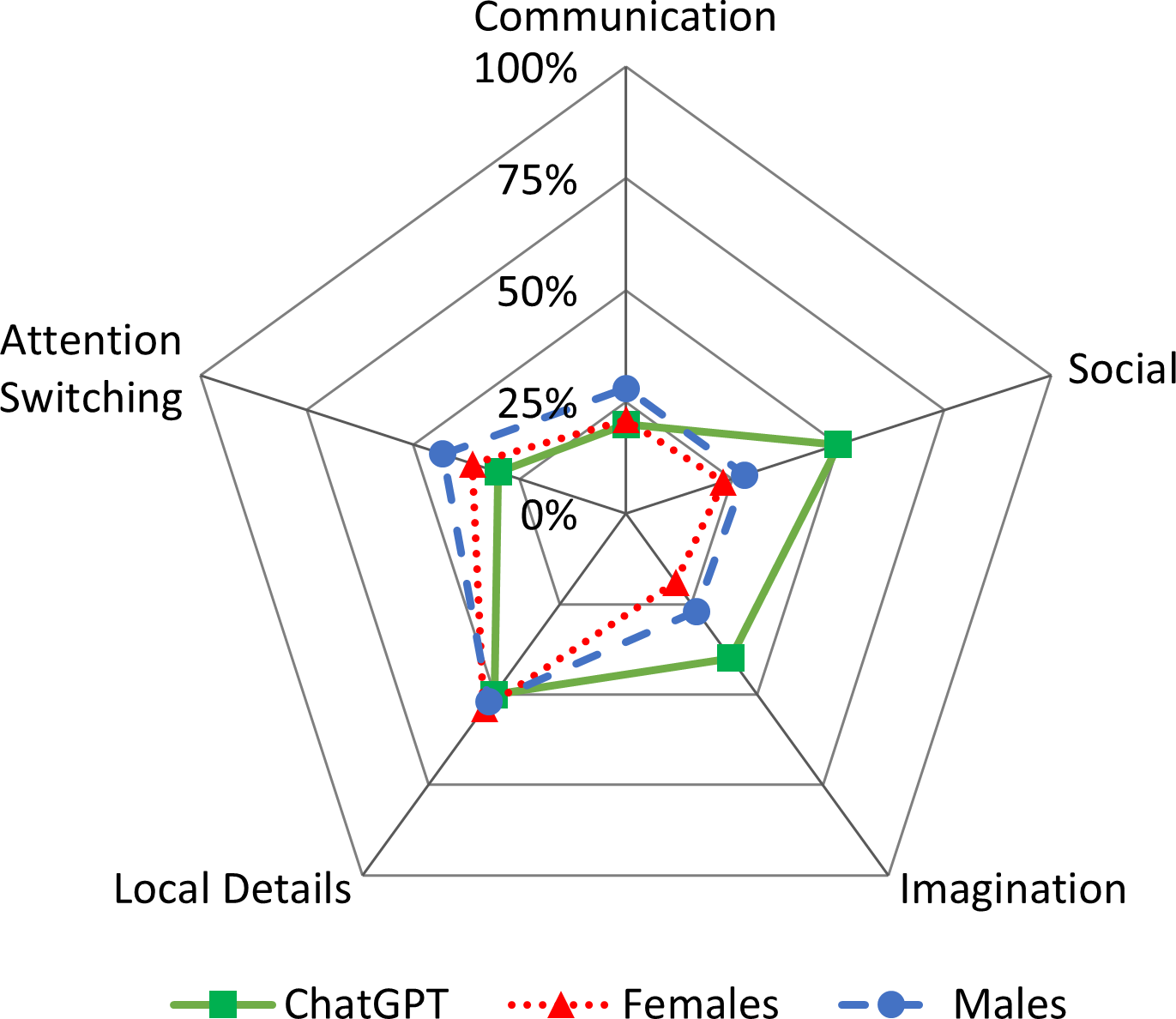}
    \caption{ChatGPT's AQ Results Compared to Males/Females.}
    \label{fig:aq_skills}
\end{figure}

Figure \ref{fig:aq_skills} illustrates how the scores are distributed amongst the respective skills compared to the average male and female scores. In the figure, the achieved scores are shown as a percentage in relation to the maximum possible score. 

\begin{table*}[ht!]
    \centering
    \begin{tabular}{lrrrrcc}
    \toprule
        ~ & Total & \multicolumn{2}{c}{Mean (SD)}  &    \\ 
        ~ & ChatGPT &  Males &  Females &  Range  & $\Delta_{ChatGPT:Males}$ & $\Delta_{ChatGPT:Females}$\\ \midrule
        IRI -- Fantasy & 17 & 15.73 (5.60) & 18.75 (5.17)  & 0-28 & ~8\% higher & 10\% higher \\ 
        ~~~~~-- Perspective Taking & 16  & 16.78 (4.72) & 17.96 (4.85) & 0-28 & ~5\% lower & 12\% lower\\ 
        ~~~~~-- Empathic Concern & 11  & 19.04 (4.21) & 21.67 (3.83) & 0-28 & 73\% lower & 97\% lower \\ 
        ~~~~~-- Personal Distress & 9 & 9.46 (4.55) & 12.28 (5.01) & 0-28 & ~5\% lower & 36\% lower \\ 
        EQ & 34 & 41.8 (11.7) & 47.2 (10.2) & 0-80 & 23\% lower & 39\% lower   \\ 
        TEQ & 41 & 43.63 (7.93) & 48.33 (6.90) & 0-64 & ~6\% lower & 18\% lower \\ 
        PES -- General Cognitive Empathy & 22  & 37.6 (7.13) & 38.4(7.06) &  0-50 & 71\% lower & 75\% lower \\ 
        ~~~~~~-- Positive Affective Empathy & 11 & 15.4 (4.05) & 16.1 (3.94) &  0-25 & 40\% lower & 46\% lower  \\ 
        ~~~~~~-- Negative Affective Empathy & 7 & 11.1 (3.39) & 12.4 (3.71) &  0-25 & 59\% lower & 77\% lower \\
        ~~~~~~-- General Empathy & 40 & 64.1 (10.92) & 66.9 (11.27) &  0-100 & 60\% lower & 67\% lower \\
        AQ & 19 & 17.8 (6.8) & 15.4 (5.7) & 0-50 & ~7\% higher & 23\% higher  \\ 
    \bottomrule
    \end{tabular}
    \caption{ChatGPT's Empathy Scores Compared to the Mean Score of Males and Females. (Note: In contrast to the scores of the other questionnaires, a higher AQ score refers to a lower level of empathy.)}
    \label{tab:summary}
\end{table*}

The scores for ChatGPT show that, especially for \textit{social skills}, the score is higher than for healthy adults, which is in line with the score for empathic concern of the IRI. This can be seen as an indicator that ChatGPT has difficulties fully connecting to other people's feelings and feeling concerned or compassionate for them. Moreover, \textit{imagination skills} are worse than the average for healthy humans. If it comes to \textit{attention switching}, ChatGPT performs considerably well, while the skill to focus on details and the \textit{communicative skills} of ChatGPT is quite similar to those of healthy humans.

\section{Conclusion and Future Work}
\label{sec:conclusion}
In the studies presented, we investigated the empathic capabilities of ChatGPT. 
In our first study, we demonstrated that ChatGPT is able to rephrase a sentence to express a particular emotion with an accuracy of 91.7\%. This shows that ChatGPT has the potential to be used as a tool for expressing emotions on demand which can help in the interaction with humans---be it in a learning environment or when being used as a source of everyday companionship. In a second study, we additionally demonstrated that ChatGPT can generate parallel emotional responses with 70.7\% accuracy, meaning that it is able to respond with the same emotion as the initial prompt in many cases. Furthermore, our results show that ChatGPT has a strong tendency to reply with \textit{joy}. 
In our last study, we used five questionnaires to test the empathic capabilities of ChatGPT. The questionnaires indicated that ChatGPT is able to interpret the emotions of others and take their perspective but still has some difficulties showing a higher level of empathy compared to healthy humans. All scores from the questionnaires in comparison to healthy males and females are summarized in Table~\ref{tab:summary}. 
 While \cite{li2022gpt} concluded that GPT-3 shows a significant lack of empathy based on the psychopathy section of the SD-3 \cite{jones2014introducing} questionnaire, in our empathy-focused studies we demonstrated that ChatGPT expresses empathy in several aspects.

With our research, we show a possible way to proceed with analyzing chatbots in the future. Further research should focus on developing more sophisticated models that can more accurately grasp the emotional context of a conversation, as well as on the development of methods to measure the emotional capabilities of a chatbot. In addition, studies should be conducted to explore how ChatGPT can be used as a tool to support people more compassionately. Finally, it is important to consider the ethical implications of using chatbots such as ChatGPT. This is particularly important because they often interact with people who may not be aware that they are interacting with a computer program. Developing methods for assessing the ethical implications of using chatbots can help ensure that they are used ethically and that potential harm is minimized.



\section*{Ethical Impact Statement}




Individuals were asked to label the text data we collected for our data annotation. The participants who supported us were not dependent on the authors and participated voluntarily and free of charge. There was no conflict of interest between the supporters and the authors. For privacy reasons, the names of the supporters are not disclosed.

The collected corpus is made freely available to the community. The collected text data of the corpus are extracts from the \textit{EmpatheticDialogues} dataset from Facebook Research~\cite{rashkin2019towards} and produced by ChatGPT. Usually, ChatGPT produces text appropriate for the general public, but it cannot be ruled out that the content is not suitable for everyone. The text data contains emotional sentences in various forms. But this is the essence of a corpus that can be used to evaluate a chatbot's output realistically.


\bibliographystyle{IEEEtran}
\bibliography{newbib}

\end{document}